\pdfoutput=1

\documentclass[11pt]{article}

\usepackage[]{emnlp2021}

\usepackage{times}
\usepackage{latexsym}
\usepackage{amsmath}
\usepackage{microtype}
\usepackage{graphicx}
\usepackage{subfigure}
\usepackage{booktabs} 
\usepackage{enumitem}

\usepackage[T1]{fontenc}

\usepackage[utf8]{inputenc}

\usepackage{microtype}

\usepackage{xurl}

\title{Neural Unification for Logic Reasoning over Natural Language}

\author{Gabriele Picco\thanks{\hspace{2pt} Equal contribution.} \\ {\bf Hoang Thanh Lam\footnotemark[1] } \\ {\bf Marco Luca Sbodio} \\ {\bf Vanessa Lopez Garcia} \\
  IBM Research / Europe \\
  \texttt{gabriele.picco@ibm.com} \\
  \texttt{\{t.l.hoang, marco.sbodio, vanlopez\}@ie.ibm.com} \\}

\begin{document}
\maketitle
\begin{abstract}

Automated Theorem Proving (ATP) deals with the development of computer programs being able to show that some conjectures (queries) are a logical consequence of a set of axioms (facts and rules).
There exists several successful ATPs where conjectures and axioms are formally provided (e.g. formalised as First Order Logic formulas).
Recent approaches, such as \cite{clark2020transformers}, have proposed transformer-based architectures for deriving conjectures given axioms expressed in natural language (English).
The conjecture is verified through a binary text classifier, where the transformers model is trained to predict the truth value of a conjecture given the axioms.
The RuleTaker approach of \cite{clark2020transformers} achieves appealing results both in terms of accuracy and in the ability to generalize, showing that when the
model is trained with deep enough queries (at least 3 inference steps), the transformers are able to correctly answer the majority of queries (97.6\%) that require up to 5 inference steps.
In this work we propose a new architecture, namely the \emph{Neural Unifier}, and a relative training procedure, which achieves state-of-the-art results in term of generalisation, 
showing that mimicking a well-known inference procedure, the \emph{backward chaining}, it is possible to answer deep queries even when the model is trained only on
shallow ones. 
The approach is demonstrated in experiments using a diverse set of benchmark data. The source code is available at this location\footnote{\url{https://github.com/IBM/Neural_Unification_for_Logic_Reasoning_over_Language}
}.

\end{abstract}

\section{Introduction}

Automated Theorem Proving (ATP) deals with the development of computer programs being able to show that some conjectures (queries) are a logical consequence of a set of axioms (facts and rules) \cite{data-exchange}. This problem has wide applications in many domains, including problem solving \cite{green1981application} and question answering \cite{maccartney2007natural,furbach2010application,hermann2015teaching,clark2020transformers}, and is being actively studied, an extensive reference can be found in \citet{10.1145/12808.12833} and \citet{nawaz2019survey}. Recent approaches, such as RuleTaker \cite{clark2020transformers}, uses transformers \cite{DBLP:journals/corr/VaswaniSPUJGKP17} as automated theorem prover over queries, facts and rules expressed in natural language (English). The theorem proving problem is translated into a binary text classification problem, where the transformers model is trained to predict the truth value (True/False) of a textual query $q$ given an input knowledge base $\kappa$ consisting of textual facts and rules.

\begin{figure*}[h!]
\center{\includegraphics[width=1\textwidth]
{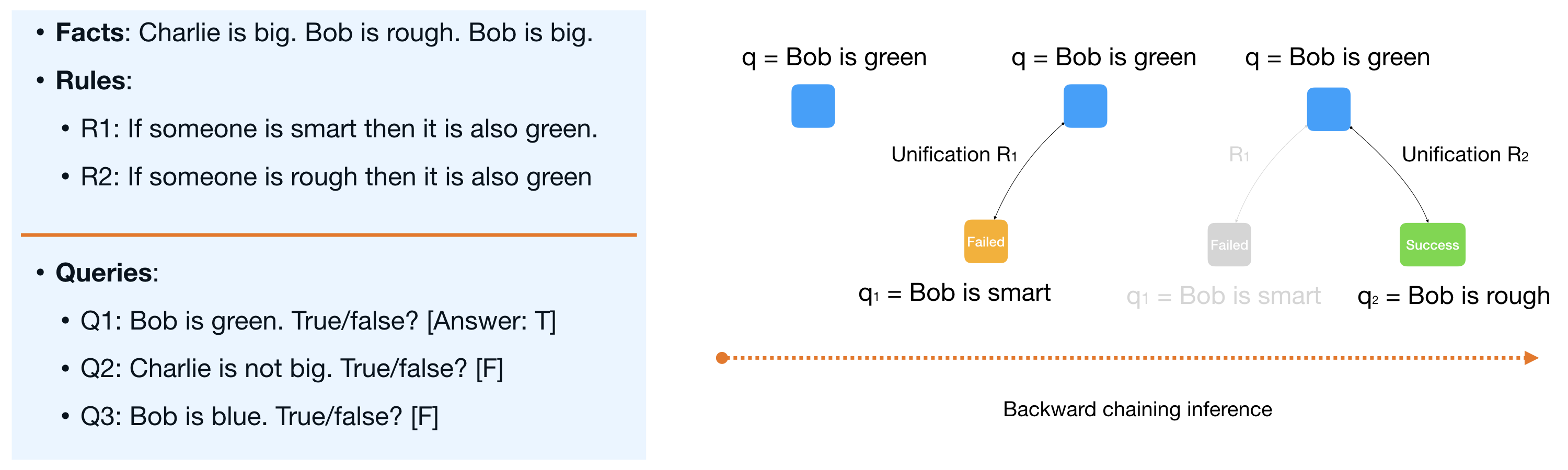}}
\caption{\label{fig:toys} A machine reasoning problem with an informally represented knowledge and query, and an example of backward chaining inference to prove a query statement: ``Bob is green."}
\end{figure*}

This class of ATP is especially interesting since it does not require the explicit translation of axioms and conjecture to formal logical (e.g First Order Logic) or probabilistic rules, making it possible to reason on knowledge expressed verbatim. Furthermore, these models do not specify an explicit reasoning procedure, but learn to implicitly demonstrate a query from example instances during the learning phase.  Figure \ref{fig:toys} shows an example of an instance of the logic reasoning problem in natural language illustrated in \citet{clark2020transformers}. 

The results reported in \citet{clark2020transformers} demonstrated that state-of-the-art pretrained language models, such as ROBERTA \cite{devlin2018bert} or BERT \cite{liu2019roberta}, can be fine-tuned with labeled data to achieve appealing results both in terms of accuracy and in the ability to generalize, showing that when the
model is trained with deep enough queries (at least 3 inference steps), the transformers are able to correctly answer the majority of queries (97.6\%) that require up to 5 inference steps.  This interesting result holds not only for training and test data in the same domain, but also for zero-shot testing on texts in other domains. 

In this paper we propose an architecture that is able to answer  deep queries (having large inference depths) even when it is trained only with shallow queries at depth 1 or 2. 
Our main assumption is that by inducing a neural network to mimic some elements of an explicit general reasoning procedure, e.g. the \textit{backward chaining},  we can increase the ability of the model to generalize. In particular we focus on mimicking the iterative process in which, at each step, a query is simplified by unifying it with an existing rule to create a new but simpler query for further checking \cite{BaaderS01}. 
In a unification step, when the query matches with the consequent (\textit{Then} clause) of a rule, the antecedent (\textit{If} clause) of the rule is combined with that query via symbolic substitution to create a new query. For example, for the query ``Bob is green" shown in Figure \ref{fig:toys}, the following steps lead to the answer (proof):

\begin{itemize}[leftmargin=*]
    \item Fact checking step 0: No fact in our knowledge base matches with the query ``Bob is green"
    \item Unification step 1: Given ``Bob is green" and the  rule: ``If someone is smart then it is also green.", a new query is created ``Bob is smart"
    \item Fact checking step 2: No fact matches with the new query ``Bob is smart"  
    \item Unification step 3: Given that ``Bob is green" and the  rule: ``If someone is rough then it is also green", a new query ``Bob is rough" is created
    \item Fact checking step 4: ``Bob is rough" matches with a fact in the knowledge base, the proof completes and returns the answer: ``Bob is green" is a true statement.
\end{itemize}

As we can see in the given example, the query ``Bob is green" is simplified iteratively with the help of the unification steps and is transformed into a factual query ``Bob is rough", which is then checked by the fact-checking step via a simple look-up in the knowledge base. These sequences of inference steps are the basis of the famous backward chaining inference in formal logic \cite{russell2002artificial} illustrated in Figure \ref{fig:toys}. 

The main building blocks of such inference methods are the unification and the fact checking algorithms. While backward chaining inference with formal representation can be formulated as a tree search problem \cite{russell2002artificial}, emulating these algorithms for textual input data using neural networks is still an open research problem, mainly due to the ambiguity in mapping entities and relations expressed in natural language to corresponding mentioning entities and relations in free-text knowledge bases.

In the following sections we describe the Neural Unifier architecture, that mimics the unification and the fact checking algorithms, in order to improve generalisation on answering deep queries. We test our approaches with publicly available datasets where the Neural Unifier is trained with depth-1 or depth-2 queries, and demonstrate that it can answer queries at higher depths with high accuracy (up to five inference steps). In particular, the proposed approach achieved state-of-the-art results in these benchmark datasets and outperformed the state-of-the-art algorithms with a significant margin. 
\section{Preliminaries and problem definition}
\label{ss:problem definition}
This section provides a formal problem definition and introduces the main intuition of the approach used for mimicking the backward chaining.

The backward chaining algorithm, described extensively in \cite{russell2002artificial}, is one of the most used algorithms for reasoning with inference rules: it is based on a depth-first strategy to explore the search space, and it generates a proof including a sequence of unification and fact checking operations. An example of execution of this algorithm is shown in Figure \ref{fig:toys}.

Let $q$  denote a the query (conjecture) and $\kappa$ be the knowledge base that consists of a set of rules $R$ and a set of facts $F$; 
In this work we consider all $q$, $R$, and $F$ expressed in natural language, where queries, facts and rules can be very simple lexicalizations of logical formulas (e.g, Figure \ref{fig:toys}); or they can be paraphrased in a more creative way. Experiments with both simple lexicalizations and paraphrases are reported in Section \ref{ss:inference-synth} and Section \ref{ss:paraphresed} respectively. Let  $q_n$ denote as  \textit{depth-n} query that requires at least $n$ inference steps in order to provide an answer. For example, query 1 in Figure \ref{fig:toys} is a depth-1 query because it requires a unification with $R_2$, plus a fact checking step (shown as the success path in Figure \ref{fig:toys}).

The fact checking function denoted as $f$ can now be formalised as the operation that takes as input a depth-0 query $q_0$ and returns True if  the corresponding fact is present in the given $\kappa$, and False otherwise:

\[ f(q_0,\kappa) = \begin{cases} 
      True & q_0 \in \kappa \\
      False & q_0 \not \in \kappa \\
   \end{cases}
\]

The unification operation denoted as $u$ can be formalised as the operation that takes as input a query $q_n$ and the set of facts and rules $\kappa$ and provides as output a simpler query $q_{n-1}$ at depth-(n-1):

\[ u(q_n,\kappa) = q_{n-1} \]

while the application of $k$ unification steps consecutively is denoted as $ u^k(q_n,\kappa) = q_{n-k} $, and for the special case where $n = k$ we have that: 

\[ u^n(q_n,\kappa) =  q_0 \]

Let now assume the existence of a perfect unification operator denoted as $u_*$, that is when it explores the search tree defined by backward chaining, always chooses the branch corresponding to the optimal path (where an optimal path for a query $q_n$ is a series of unification operations leading to a query $q_0$ with the same truth value of $q_n$  in $n$ steps). Considering a relaxed version of the problem where the queries do not require closed or open word assumption to prove, the truth value of a query $q_n$ can therefore be found with $n$ unification steps plus a fact checking operation: $f(u_*^{n}(q_n,\kappa),\kappa)$. 

With a symbolic representation, the unification and the fact checking operations can be done via explicit mathematical transformations. However, when the input is represented in natural language without an explicit structure, it requires machines to learn these tasks by examples under the presence of language ambiguity. In this work, we propose a neural network architecture called Neural Unifier (NU) aiming at learning to approximate the function $f(u_*^{n}(q_n,\kappa),\kappa)$ with input expressed in natural language.  Details about our approach are discussed in the next section.
\begin{figure*}[thb]
\center{\includegraphics[width=1.0\textwidth]
{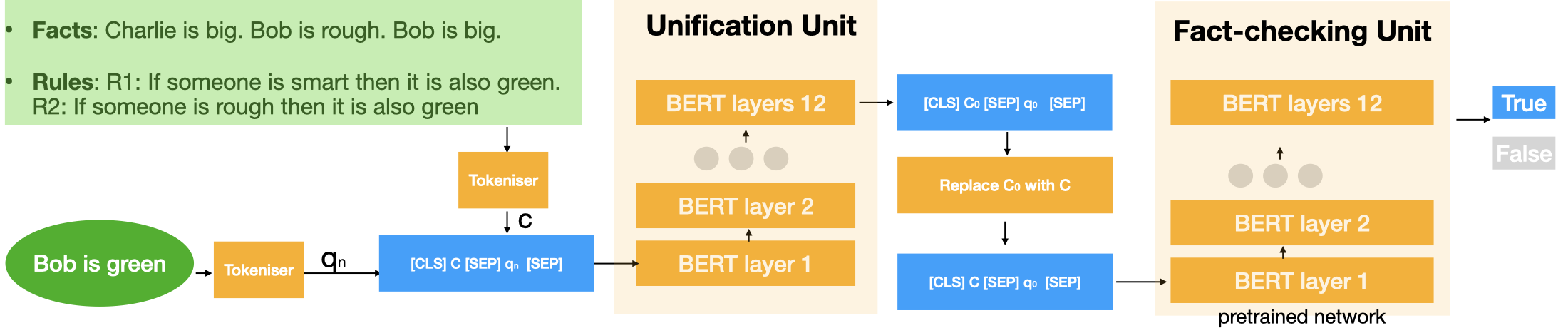}}
\caption{\label{fig:nu} A Neural Unification network consists of a pretrained fact-checking unit and a unification unit. The  fact-checking unit is a model trained to check whether an embedding vector of depth-0 query $q_0$ is true/false given a knowledge base embedding vector $C$. The unification unit takes an embedding vector $q_n$ of a depth-$n$ query and an embedding vector $C$ of the knowledge base  as an input. It transforms the  input vector $q_n$, thanks to the multi-attention layers, to predict an embedding vector $q_0$ such that the pretrained fact look-up unit can make a correct prediction of the query's label.}
\end{figure*}
\section{Training Procedure}
\label{s:mimic}

With a slight abuse of notation, in order to simplify the discussion, when we use $q$ to denote the query, we also refer it as a notation of the embedding vector of the query, because a textual query in a neural network is represented as an embedding vector. The main idea behind the approach presented is an architecture composed of two units trained in two separate phases:

\begin{enumerate}[leftmargin=*]
\item The first unit  is the \textbf{Fact-checking Unit (FU)}: it approximates the fact checking operator $f(q_0, \kappa)$. The model is pre-trained in a supervised manner on only depth-0 queries and the related $\kappa$. After this initial training phase the FU weights are frozen and the model is used solely for predictions.
\item The second unit is the \textbf{Unification Unit (UU)}: it is trained in a second subsequent phase and the goal is to approximates the $u_*^n(q_n, \kappa)$ operator. The model is trained on depth-n queries (with $n > 0$) to produce an embedding vector $q_0$.  The embedding vector $q_0$ is then fed as an input to the pretrained FU unit whose output prediction (True or False) is used for back-propagating the error in the NU model. 
\end{enumerate}

While the first phase of training is rather intuitive, the second phase teaches the unification unit, starting from a query $q_n$, to transform it into a vector embedding   $q_0$ such that the neural fact checking unit can predict the correct truth value for the query $q_n$. Since the FU unit is pretrained to perform fact-checking tasks, the UU unit is forced to produce a query $q_0$ from a complex query $q_n$ so that the answer given by the FU unit is correct, hence mimicking the unification operations. The detailed implementation and input/output of the two units are described in detail in the following subsections.
\subsection{Neural fact checking unit}
Figure \ref{fig:nu} illustrates the core components of our proposed  Neural Unifier (NU) architecture: the Fact checking Unit (FU) and the  Unification Unit (UU). The FU component is implemented through a standard Bert transformers model \cite{bert} with a binary classification head. The inputs during the training phase are textual tuples containing the set of facts and rules $\kappa$ (concatenated in a single string), and a related depth-0 query.
The Bert tokenizer is used to transform $\kappa$ into the corresponding embedding vector denoted as $C$ ( C stands for ``context") and the depth-0 query into its embedding vector representation $q_0$. The output of the tokenization step therefore follows the format:
$[CLS] \: C \: [SEP] \: q_0 \: [SEP]$ where  $[CLS]$ and $[SEP]$ are embedding vectors of special tokens added by the BERT tokenizer to separate context and query \cite{bert}.

The transformer model is then fed with the tokenized input and fine-tuned to output \texttt{True} if the query is a correct conjecture with respect to the set of textual facts and rules provided in the input, \texttt{False} otherwise. 
Note that after the first learning phase the weights of the FU unit are frozen when the UU unit is being trained.

\subsection{Neural unification unit}
In our implementation, we also use a Bert transformers as a unification unit, with the only difference that in this case the output is an embedding vector (while the neural fact checking unit has a binary classification head and output True/False labels). The inputs for the  unification unit are textual tuples containing the set of facts and rules $\kappa$, and the related depth-n query ($n>0$).

The Bert tokenizer is used to transform $\kappa$ into the corresponding embedding vector $C$  and the depth-n query into the corresponding embedding vector $q_n$. The output of the tokenization step therefore follows the format:
$[CLS] \: C \: [SEP] \: q_n \: [SEP]$. This input is fed into the transformers to get the output (the hidden states of the transformers last layers) in the format $[CLS] \: C_0 \: [SEP] \: q_0 \: [SEP]$. Where $C_0$ and $q_0$ are the transformations of $C$ and $q_n$ via transformers respectively. 

\subsection{Wiring unification and fact checking}
Figure \ref{fig:nu} shows the complete architecture when two units are wired into one network. In particular, the output of the UU unit is  fed into the FU unit. However, in our implementation, instead of using $[CLS] \: C_0 \: [SEP] \: q_0 \: [SEP]$ as a direct input to the fact checking unit, we replace $C_0$ by the original context embedding $C$, hence the corrected input to the FU unit is $[CLS] \: C \: [SEP] \: q_0 \: [SEP]$. In doing so the  unification unit can focus on optimizing the prediction of the query embedding $q_0$ rather than trying to reconstruct the original context. We observed in experiments that this approach simplifies the learning process and helps converging faster to the optimal solutions. Detailed examples of inputs and outputs can be found in the appendix.

\section{Experiments and Results}

This section discusses experimental settings and results.

\subsection{Datasets and experimental settings}

We used three datasets provided in \cite{clark2020transformers} to validate our approach. These datasets were selected because they contains test queries that require up to 5 inference steps, where we can validate the induced generalization capability of our approach. 
The three datasets are:
\begin{itemize}[leftmargin=*]
    \item Rule reasoning data: synthetically created from a synthetic knowledge base (see \cite{clark2020transformers} for more information about the data creation process). It consists of 5 folders with depths ranging from 0 to 5. We only use depth-0 training queries in the depth-0 folder for training the fact-checking unit. Depth-1 and depth-2 training queries in the corresponding folders are used for training the unification models. The folders depth-3, depth-4 and depth-5 are used as holdout sets only for testing purposes.
    \item Paraphrased data: created from rule reasoning data where the questions, facts and rules are paraphrased by crowd-workers. Paraphrased dataset contains more complex and longer sentences, such as: \textit{"Alan is young and green, and seems to be cold and rough, but time will round him into a decent person"} (see \cite{clark2020transformers} for details). We use this dataset to test the zero shot generalisation capability to a larger variety of more natural linguistic forms.
    \item Electricity data: synthetically created from a set of rules on an electrical circuit, describing the conditions for an appliance to function. Contains queries that require up to 4 inference steps (see \cite{clark2020transformers} for details).
\end{itemize}

All datasets, with the exception of the \textit{Electricity data}, are divided into training, validation and testing sets. More details can be found in the appendix.

\subsection{Methodology}
\label{ss:methodology}

In all experiments, we used BERT (specifically \textit{bert-base-uncased}) as our backbone model both for the FU and the UU unit. 
For both training phases, described in Section \ref{s:mimic}, we used the Adam optimisation algorithm \cite{kingma2014adam} with logloss \cite{DBLP:journals/corr/Vovk15}, we set the mini-batch size as 8 to fit our GPU memory and manually fine-tuned the learning rate in the range $[10^{-6}, 10^{-2}]$, choosing the best learning rate by looking at the accuracy of the prediction in the validation set during training (0.0001). It is important to notice that the test sets used for reporting experimental results are different from the validation sets used for hyper-parameter optimization to ensure that the comparison is fair.  In all the training we used early stopping technique on the validation for avoiding overfitting, by setting the maximum number of epochs to 20. 

\subsubsection{Neural Fact-checking unit (FU)}
\label{sss:nfcu-train}

The FU unit used in the experiments is trained on 58,844 depth-0 queries (all depth-0 training queries in the depth-0 folder of \textit{Rule reasoning data}). 
The training task on the \textit{Rule reasoning data} turns out to be particularly simple, after a few epochs (around 3 using the early stopping strategy) the model is able to solve the task perfectly on the training and validation set and the report an accuracy on the set (1,6751 queries) is close to one (0.99968).
The FU also achieves an accuracy of 0.71 and 0.99 on the respective \textit{Paraphrased data} (2,968 queries) and \textit{Electricity data} (2,812 queries) test sets without any further fine-tuning (zero shot setting). 
Achieving high accuracy in the fact-checking unit is particularly important as the subsequent UU training assumes the FU prediction as ground truth to back-propagate the gradient.

\subsubsection{Neural unification unit}
\label{sss:nnu-train}

Since our work focuses on learning a general inference mechanism on shallow queries and applying it to solve queries at higher depths (up to 5 inference steps), we report the experiments on two variants of the Neural Unifier, trained on queries with a maximum depth of 2:

\begin{itemize}[leftmargin=*]
\item NU (D = 1): that is a Neural Unifier network composed of an UU unit trained on 27429 depth-1 queries (all depth-1 training queries in the depth-1 folder of \textit{Rule reasoning data}) and the previously trained FU as described in section \ref{sss:nfcu-train}
\item NU (D = 2): that is a Neural Unifier network composed of a UU unit trained on 14254 depth-2 queries (all depth-2 training queries in the depth-2 folder of \textit{Rule reasoning data}) and the previously trained FU as described in section \ref{sss:nfcu-train}
\end{itemize}

Both the models try to induce the inference mechanism, using the procedure described in section \ref{s:mimic}. NU (D = 1) reduces 1-depth queries to 0-depth equivalent vector embedding and then use the fact checker to derive the truth value of the original query. NU (D = 1) only observes queries at depth 0 and 1 during the training phase.
NU (D = 2) instead turns depth-2 queries to depth-0 equivalent vector embedding and then uses the fact checker to prove the conjecture. Therefore NU (D = 2) only observes queries at depth 0 and 2 during the training phase. In the following sections, the results of the two models will be reported (trained with the settings described in subsection \ref{ss:methodology}), focusing on the generalization capacity to queries at depths not observed during training.

\subsection{Inference on queries deeper than those observed at training time}
\label{ss:inference-synth}

Table \ref{comparation1-table} reports the performances of the NU $D = 1$ and NU $D = 2$ models and compares them with the state-of-the art RuleTaker approach. There are two version of RuleTaker in our experiments: RT is our implementation that uses \textit{bert-base} as backbone architecture, while RR corresponds to the implementation of \cite{clark2020transformers} with \textit{roberta-large} and is reported for completeness when the results are available in the original paper. 

As can be seen in this experiment, the NU models with $D=1$ and $D=2$ accurately answer queries at unseen depths, and consistently outperforms the state-of-the-art approaches on those depths. The significant result is particularly evident for NU $D = 2$ over depth-5 queries. More interesting is that  the model NU $D = 2$  not only learns to transform depth-2 queries to depth-0 equivalent vector embedding, but it can reduce a $q^n$ queries with depths ranging from 3 up to 5 to depth-0 equivalent vector embedding effectively. Our hypothesis is that the transformers-based architecture, which in several applications has been shown to efficiently learn recursive tasks \cite{DBLP:journals/corr/VaswaniSPUJGKP17}, effectively approximates the unification operator $u_*^n(q_n, \kappa)$ with its multi-layer architecture described in section \ref{ss:problem definition}.

\begin{table}[h!]
\caption{Accuracy on the Rule reasoning test sets when the depths of the test queries are varied. NU $D=1$ and NU $D=2$ are compared with our implementation of the-state-of-the-art RuleTaker (RT) approach with \textit{bert-base-uncased} back-bone and the original implementation of the RuleTaker (RR) with \textit{roberta-large} back-bone pretained on the RACE dataset as reported in \citet{clark2020transformers}. }
\vskip -0.2in
\label{comparation1-table}
\begin{center}
\begin{small}
\begin{sc}
\center{\includegraphics[width=.49\textwidth]
{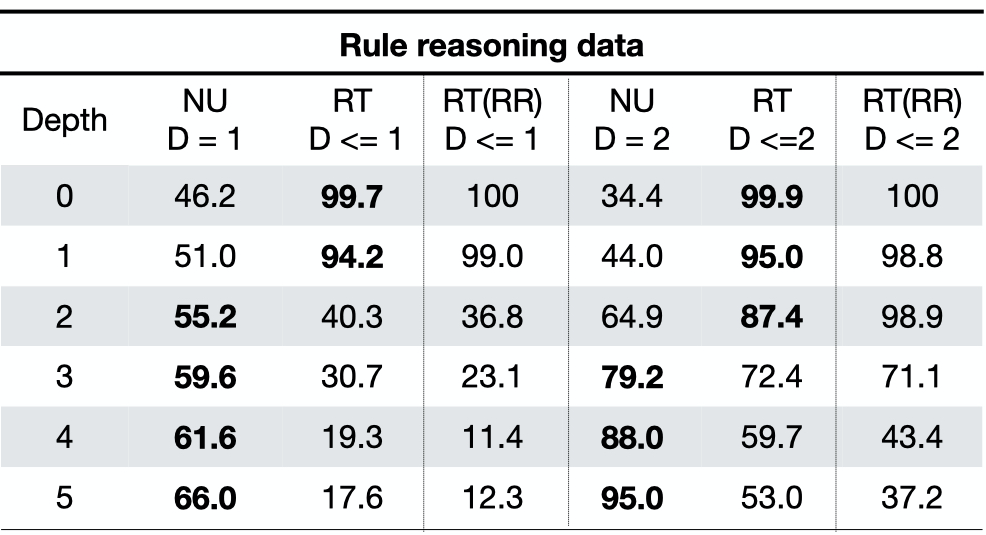}}
\end{sc}
\end{small}
\end{center}
\vskip -0.30in
\end{table}

Although our implementation uses Bert, several transformers can be used successfully while maintaining the properties analyzed, as demonstrated in \citet{clark2020transformers} and the additional experiments are reported in the appendix.

\subsection{Inference on provable queries deeper than those observed at training time}

Observing the table \ref{comparation1-table}, it may be counter-intuitive that the reported accuracy for NU (D = 1) and NU (D = 2) increases as the depth of the queries increases.
This fact can be explained by observing a bias present in the distribution of the queries that does not have a proof, and its truth value is assigned based on the closed word assumption (CWA). We call those queries CWA while the other ones, which have at least one successful proof, are called  \textit{}{provable queries} in the test set.

\begin{table}[h!]
\caption{Distribution of CWA  queries in the test data.}
\vskip -0.2in
\label{distribution-table}
\begin{center}
\begin{small}
\begin{sc}
\center{\includegraphics[width=.49\textwidth]
{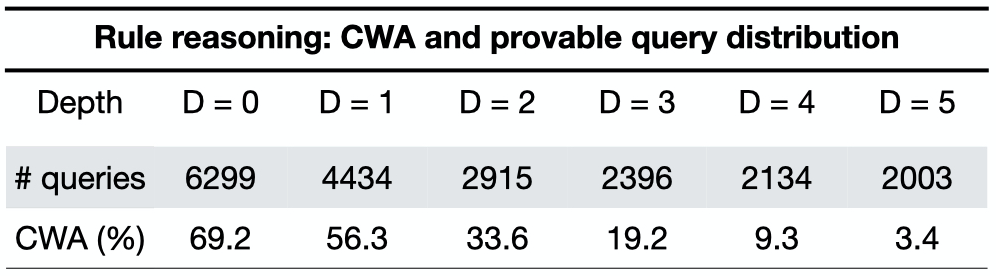}}
\end{sc}
\end{small}
\end{center}
\vskip -0.30in
\end{table}

Table \ref{distribution-table} shows the distribution of CWA and provable queries in our test data. The statistics shows an inversely proportional relationship between the number of CWA questions and the accuracy of the models reported in Table \ref{comparation1-table}, thus suggesting that NU network are especially effective on provable queries, while they do not work so well on queries that does not have a proof (CWA).

This assumption is verified by testing the models on the subset of provable queries, as reported in Table \ref{comparation2-table}.

These results show that NU (in particular NU $D = 2$) is able to answer provable queries, at all depths very accurately. Based on this observation, in the following subsections, we will demonstrate that by combining NU with RuleTaker to form an ensemble model, we are able to outperform each individual model for both CWA and provable queries.

\begin{table}[h!]
\caption{Accuracy of NU $D=1$, NU $D=2$, RT $D<=1$ and RT $D<=2$ on the provable queries in the Rule reasoning test sets (queries that does not require closed word assumption to prove the conjectures). }
\vskip -0.2in
\label{comparation2-table}
\begin{center}
\begin{small}
\begin{sc}
\center{\includegraphics[width=.49\textwidth]
{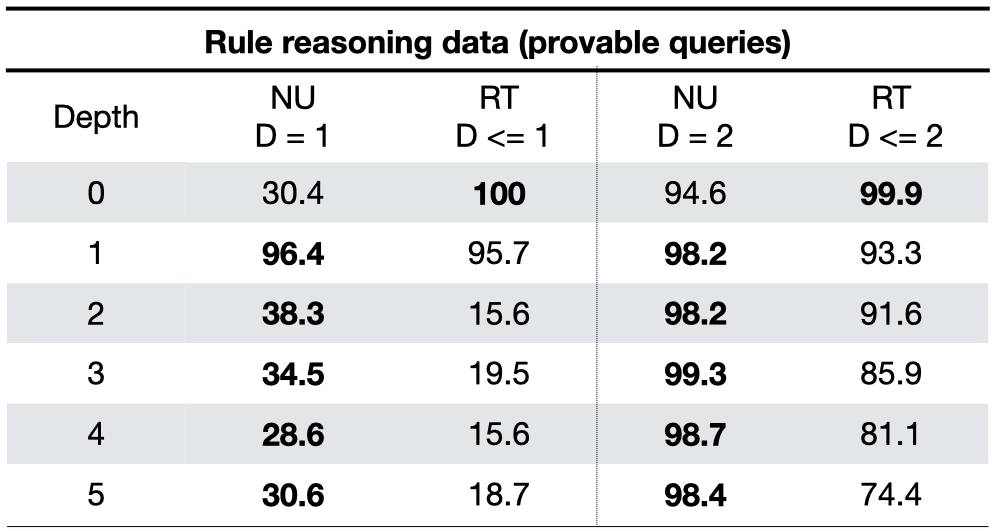}}
\end{sc}
\end{small}
\end{center}
\vskip -0.3in
\end{table}

\subsection{Zero-shot generalization}
\label{ss:paraphresed}

In order to verify the ability to generalize to deep unseen provable queries, we test the NU ($D=2$) on the provable queries in the Paraphrased and Electricity datasets, without fine-tuning the model (zero shot setting). 

The results, reported in Table \ref{comparation3-table}, shows that NU ($D=2$) can effectively outperform the current state-of-the-art by a significant margin in both accuracy and generalisation capability.

\begin{table}[h!]
\caption{Accuracy of NU $D=2$ and RT $D<=2$ on the subset of provable queries in the Paraphrased and Electricity test sets (queries that does not require closed word assumption to prove the conjectures). }
\vskip -0.2in
\label{comparation3-table}
\begin{center}
\begin{small}
\begin{sc}
\center{\includegraphics[width=.49\textwidth]
{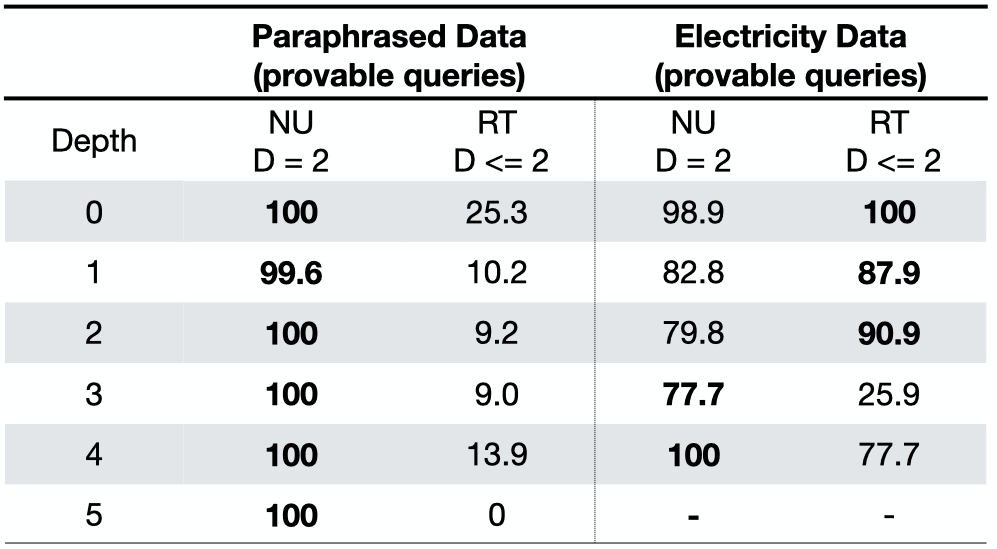}}
\end{sc}
\end{small}
\end{center}
\vskip -0.35in
\end{table}

\subsection{Weighted Ensemble methods}

While sections \ref{ss:inference-synth} and \ref{ss:paraphresed} show the effectiveness of the NU approach on provable queries, we also propose a linear weighted ensemble approach that combines the prediction of both NU and the RuleTaker. The key idea behind the ensemble method is that NU demonstrated working very well for provable queries while the RuleTaker was very good at answering CWA queries. Therefore, by combining these approaches we are able to handle both types of queries effectively.  In order to choose the weights, we tune them to optimize the accuracy on the available validation sets for each depth. The results is reported on a separate test set.

\begin{table}[h!]
\caption{Accuracy of the weighted ensemble of NU ($D=2$) and RT ($D<=2$) on the Synthetic and Paraphrased test sets. }
\vskip -0.2in
\label{ensembling-table}
\begin{center}
\begin{small}
\begin{sc}
\center{\includegraphics[width=.49\textwidth]
{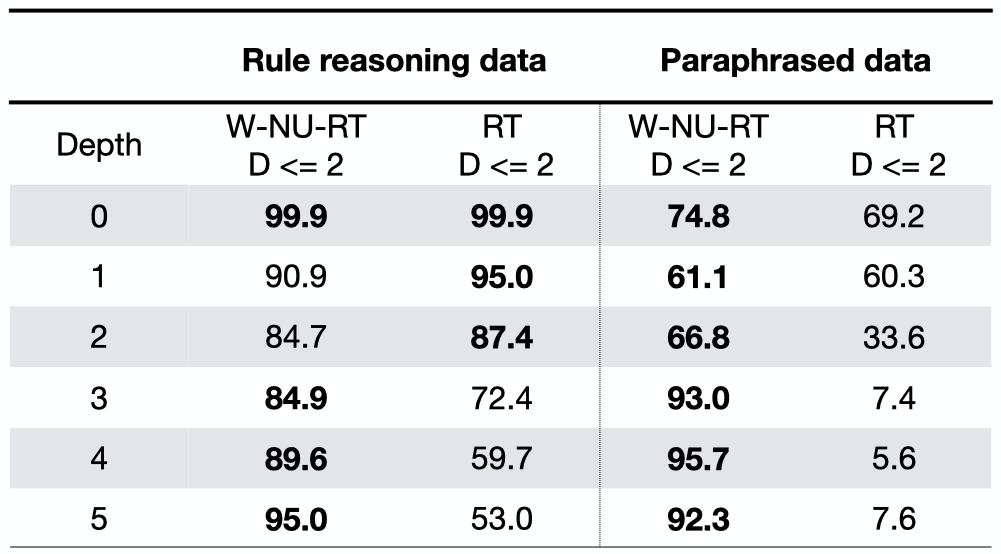}}
\end{sc}
\end{small}
\end{center}
\vskip -0.3in
\end{table}

Table \ref{ensembling-table} reports the results of the ensemble approach (W-NU-RT) on the test data. It can be seen that the weighted ensemble of NU $D = 2$ and RT $D <= 2$ effectively leverages the advantages of the two approaches to answer both CWA and provable queries at all depths effectively. The ensemble method outperforms the RuleTaker in both datasets and at most depths.

\subsection{Significance tests}

We highlight below some significant results obtained with statistical tests:

\begin{itemize}[leftmargin=*]
    \item Table \ref{comparation2-table}: the model NU $D = 2$ has significantly better (p-value $0.020$ computed with randomization test) results than RT $D <= 2$ (previous state-of-the-art) on provable queries.
    \item Table \ref{comparation3-table}: the model NU $D = 2$ has significantly better (p-value 0.002 computed with randomization test) results than RT $D <= 2$ (previous state-of-the-art) on paraphrased provable queries.
     \item Table \ref{ensembling-table}: W-NU-RT has significantly better (p-value 0.008 computed with randomization test) results than RT on paraphrased data (columns 3 and 4 of table \ref{ensembling-table}) without compromising its performance on rule-reasoning data (columns 1 and 2 of table \ref{ensembling-table}).
\end{itemize}

We also highlight an aspect not evident from the statistical tests, our proposed model outperforms in all experiments, at depth 3,4,5 not seen during training, previous state-of-the-art.

\section{Related work}

While our work is, to the best of our knowledge, the first proposed architecture that emulates backward chaining inference over rule sets and facts expressed in natural language, there are several methods which explore this research area.
\cite{clark2020transformers} introduce the use of transformers to reason over explicitly stated rule sets expressed in natural language; their approach show that transformers are able to solve the problem with high accuracy when the neural network is trained with sufficiently deep reasoning paths, without imposing any structure on the neural reasoning.
Our main contribution with respect to \cite{clark2020transformers} consists in emulating, through a neural network, a general reasoning mechanism inspired by the backward chaining algorithm used in formal logic programming. Also, our method demonstrate better accuracy for high depth queries, even when trained only with shallow queries.

\cite{saha2020prover} and \cite{tafjord2020proofwriter} modify the \cite{clark2020transformers} approach in order to generate proof together with the predicted truth value, these methods however require the explicit knowledge of the proof during the training phase.

Furthermore, our work is substantially different from the methods that focus on an initial translation of knowledge expressed in textual form to a formal specification, with the aim of applying classic reasoning algorithms, such as the architecture proposed in \cite{Singh2020} for translating text into first order logic formulas. Our work is also different from \cite{NIPS2013_5028}: the authors present a neural network suitable for reasoning over relationships between two entities of a knowledge base, focusing specifically on predicting additional true facts using only vector representations of existing entities in the knowledge base.
Other approaches have combined neural and symbolic reasoning methods. One notable example is the Neural Theorem Proving (NTP) presented in \cite{Rocktaschel2017}. The authors propose an end-to-end differentiable prover, operating on symbolic representations, for automated completion of a knowledge base: they recursively construct neural networks to prove queries on the knowledge base by following Prolog’s backward chaining algorithm. Additionally, they introduce a differentiable unification operation between vector representations of symbols.
\cite{Minerv} describes an NTP capable of jointly reasoning over KBs and natural language corpora. Although the method is versatile, explicit mapping to entities in the KB is required.
Other relevant methods implement forms of neural reasoning starting from a formal knowledge base, including \cite{serafini2016logic}, \cite{Guha2014}, and \cite{NLM}, or starting from an ontology (which usually define not just the predicates, but also rules) \cite{HoheneckerL17}. Conversely, we focus on using transformers both as a fact look-up model (over a knowledge base expressed in natural language), and as a unification unit for transforming queries, which may require many steps of inference, into factual queries that can be answered with the fact look-up model. Some early work on simulating the first-order algorithm of unification using neural networks is presented in \cite{komendantskaya2011unification}. The author shows how error-correction learning algorithm can be used for the purposes of unification. However, this work considers a version of the problem where the knowledge is represented using a formal first order logic language, and uses an explicit mapping of each symbol of the language into a input vector. Similarly to our work, \cite{Weber2019NLPrologRW} approach the problem of reasoning over natural language emulating unification. They present a model combining neural networks with logic programming for solving multi-hop reasoning tasks over natural language. In the proposed approach the authors extend a Prolog prover to use a similarity function over pretrained sentence encoders. A substantial difference with respect to our work is that \cite{Weber2019NLPrologRW} approach requires the transformation of natural language text into triples (by using co-occurrences of named entities), and then embedding representations of the symbols in a triple using an encoder.

In this paper, we take inspiration from the method presented in \cite{DBLP:conf/iclr/HudsonM18}, a recurrent cell that simulates a reasoning step, although the architecture is specially designed for performing visual reasoning given a textual query.

Our goal is quite different from answering complex multi-hop questions using a corpus of documents as a virtual knowledge base as proposed in the recent work by \cite{DBLP:conf/iclr/DhingraZBNSC20}, which requires selecting spans from paragraphs of texts. Our work can be described as the formalization of a model and a training process that leads the neural network emulate a backward chaining inference process for answering deep queries.

\section{Conclusion and Future Work}

In this paper we have shown (in a limited, but not trivial setting) that machines can be trained to perform deep reasoning over language, even if trained only on shallow reasoning.
The presented approach performs inference without the need for a translation phase from natural language to a formal specification, and it obtains high accuracy on the datasets considered.
Furthermore, with a particular learning architecture that brings learning closer to a deductive argument form, help improving the ability to generalize to deep queries.
Although this work is a step in the direction of combining the ability of neural networks to emulate reasoning on non-formal data with the explanatory power of a formal demonstration procedure, further work is needed to fill the gaps.
In an ideal situation, a machine should perform  $n$ inference steps (with explicit reference to the parts of the text concerned) to answer a query with depth n.
Moreover, the reasoning procedure should be able to reason on any possible textual expression of rules or facts, excluding ambiguous and irrelevant information. With further advances, we may potentially be able to:

\begin{itemize}[leftmargin=*]
    \item Understand if there exists a relationship between the output embedding of the Neural Unification unit and an interpretable representation.
  \item Apply the Neural Unifier approach on other types of logical inference (e.g. inductive and abductive) on a different type of datasets, for example with an open word assumption.
  \item Complement the answer to a deep query, produced by our Neural Unification unit, with a (possibly approximate) formal and human interpretable proof of the answer and identify the parts of the text involved in the $n$ inference steps that led to a conclusion. One approach could be to modify the architecture by explicitly requesting evidence as input, in line with the ideas presented in \cite{saha2020prover} and \cite{tafjord2020proofwriter}.
\end{itemize}


\bibliography{anthology,bibliography}

\begin{thebibliography}{31}
\expandafter\ifx\csname natexlab\endcsname\relax\def\natexlab#1{#1}\fi

\bibitem[{Baader and Snyder(2001)}]{BaaderS01}
Franz Baader and Wayne Snyder. 2001.
\newblock Unification theory.
\newblock In John~Alan Robinson and Andrei Voronkov, editors, \emph{Handbook of
  Automated Reasoning (in 2 volumes)}, pages 445--532. Elsevier and MIT Press.

\bibitem[{Clark et~al.(2020)Clark, Tafjord, and
  Richardson}]{clark2020transformers}
Peter Clark, Oyvind Tafjord, and Kyle Richardson. 2020.
\newblock \href {https://doi.org/10.24963/ijcai.2020/537} {Transformers as soft
  reasoners over language}.
\newblock In \emph{Proceedings of the Twenty-Ninth International Joint
  Conference on Artificial Intelligence, {IJCAI} 2020}, pages 3882--3890.
  ijcai.org.

\bibitem[{Devlin et~al.(2019{\natexlab{a}})Devlin, Chang, Lee, and
  Toutanova}]{devlin2018bert}
Jacob Devlin, Ming-Wei Chang, Kenton Lee, and Kristina Toutanova.
  2019{\natexlab{a}}.
\newblock \href {http://arxiv.org/abs/arXiv:1704.04683} {Bert: Pre-training of
  deep bidirectional transformers for language understanding}.
\newblock In \emph{Proc. of 2019 Conference of the North {A}merican Chapter of
  the Association for Computational Linguistics: Human Language Technologies},
  volume~1, pages 4171--4186.

\bibitem[{Devlin et~al.(2019{\natexlab{b}})Devlin, Chang, Lee, and
  Toutanova}]{bert}
Jacob Devlin, Ming-Wei Chang, Kenton Lee, and Kristina Toutanova.
  2019{\natexlab{b}}.
\newblock \href {https://doi.org/10.18653/v1/N19-1423} {{BERT}: Pre-training of
  deep bidirectional transformers for language understanding}.
\newblock In \emph{Proceedings of the 2019 Conference of the North {A}merican
  Chapter of the Association for Computational Linguistics: Human Language
  Technologies, Volume 1 (Long and Short Papers)}, pages 4171--4186,
  Minneapolis, Minnesota. Association for Computational Linguistics.

\bibitem[{Dhingra et~al.(2020)Dhingra, Zaheer, Balachandran, Neubig,
  Salakhutdinov, and Cohen}]{DBLP:conf/iclr/DhingraZBNSC20}
Bhuwan Dhingra, Manzil Zaheer, Vidhisha Balachandran, Graham Neubig, Ruslan
  Salakhutdinov, and William~W. Cohen. 2020.
\newblock \href {https://openreview.net/forum?id=SJxstlHFPH} {Differentiable
  reasoning over a virtual knowledge base}.
\newblock In \emph{8th International Conference on Learning Representations,
  {ICLR} 2020, Addis Ababa, Ethiopia, April 26-30, 2020}. OpenReview.net.

\bibitem[{Dong et~al.(2019)Dong, Mao, Lin, Wang, Li, and Zhou}]{NLM}
Honghua Dong, Jiayuan Mao, Tian Lin, Chong Wang, Lihong Li, and Denny Zhou.
  2019.
\newblock \href {https://openreview.net/forum?id=B1xY-hRctX} {Neural logic
  machines}.
\newblock In \emph{7th International Conference on Learning Representations,
  {ICLR} 2019, New Orleans, LA, USA, May 6-9, 2019}. OpenReview.net.

\bibitem[{Furbach et~al.(2010)Furbach, Gl{\"o}ckner, and
  Pelzer}]{furbach2010application}
Ulrich Furbach, Ingo Gl{\"o}ckner, and Bj{\"o}rn Pelzer. 2010.
\newblock An application of automated reasoning in natural language question
  answering.
\newblock \emph{Ai Communications}, 23(2-3):241--265.

\bibitem[{Green(1981)}]{green1981application}
Cordell Green. 1981.
\newblock Application of theorem proving to problem solving.
\newblock In \emph{Readings in Artificial Intelligence}, pages 202--222.
  Elsevier.

\bibitem[{Guha(2014)}]{Guha2014}
Ramanathan Guha. 2014.
\newblock \href {http://arxiv.org/abs/1410.5859} {{Towards a Model Theory for
  Distributed Representations}}.
\newblock \emph{AAAI Spring Symposium - Technical Report}, SS-15-03:22--26.

\bibitem[{Hermann et~al.(2015)Hermann, Kocisk{\'{y}}, Grefenstette, Espeholt,
  Kay, Suleyman, and Blunsom}]{hermann2015teaching}
Karl~Moritz Hermann, Tom{\'{a}}s Kocisk{\'{y}}, Edward Grefenstette, Lasse
  Espeholt, Will Kay, Mustafa Suleyman, and Phil Blunsom. 2015.
\newblock \href
  {https://proceedings.neurips.cc/paper/2015/hash/afdec7005cc9f14302cd0474fd0f3c96-Abstract.html}
  {Teaching machines to read and comprehend}.
\newblock In \emph{Advances in Neural Information Processing Systems 28: Annual
  Conference on Neural Information Processing Systems 2015, December 7-12,
  2015, Montreal, Quebec, Canada}, pages 1693--1701.

\bibitem[{Hohenecker and Lukasiewicz(2017)}]{HoheneckerL17}
Patrick Hohenecker and Thomas Lukasiewicz. 2017.
\newblock \href {http://arxiv.org/abs/1705.10342} {Deep learning for ontology
  reasoning}.
\newblock \emph{CoRR}, abs/1705.10342.

\bibitem[{Hudson and Manning(2018)}]{DBLP:conf/iclr/HudsonM18}
Drew~A. Hudson and Christopher~D. Manning. 2018.
\newblock \href {https://openreview.net/forum?id=S1Euwz-Rb} {Compositional
  attention networks for machine reasoning}.
\newblock In \emph{6th International Conference on Learning Representations,
  {ICLR} 2018, Vancouver, BC, Canada, April 30 - May 3, 2018, Conference Track
  Proceedings}. OpenReview.net.

\bibitem[{Kingma and Ba(2015)}]{kingma2014adam}
Diederik~P. Kingma and Jimmy Ba. 2015.
\newblock \href {http://arxiv.org/abs/1412.6980} {Adam: {A} method for
  stochastic optimization}.
\newblock In \emph{3rd International Conference on Learning Representations,
  {ICLR} 2015, San Diego, CA, USA, May 7-9, 2015, Conference Track
  Proceedings}.

\bibitem[{Komendantskaya(2011)}]{komendantskaya2011unification}
Ekaterina Komendantskaya. 2011.
\newblock Unification neural networks: unification by error-correction
  learning.
\newblock \emph{Logic Journal of the IGPL}, 19(6):821--847.

\bibitem[{Liu et~al.(2019)Liu, Ott, Goyal, Du, Joshi, Chen, Levy, Lewis,
  Zettlemoyer, and Stoyanov}]{liu2019roberta}
Yinhan Liu, Myle Ott, Naman Goyal, Jingfei Du, Mandar Joshi, Danqi Chen, Omer
  Levy, Mike Lewis, Luke Zettlemoyer, and Veselin Stoyanov. 2019.
\newblock \href {http://arxiv.org/abs/1907.11692} {Roberta: {A} robustly
  optimized {BERT} pretraining approach}.
\newblock \emph{CoRR}, abs/1907.11692.

\bibitem[{Loveland(1986)}]{10.1145/12808.12833}
D~W Loveland. 1986.
\newblock \href {https://doi.org/10.1145/12808.12833} {Automated theorem
  proving: Mapping logic into ai}.
\newblock In \emph{Proceedings of the ACM SIGART International Symposium on
  Methodologies for Intelligent Systems}, ISMIS '86, page 214–229, New York,
  NY, USA. Association for Computing Machinery.

\bibitem[{MacCartney and Manning(2007)}]{maccartney2007natural}
Bill MacCartney and Christopher~D. Manning. 2007.
\newblock \href {https://www.aclweb.org/anthology/W07-1431} {Natural logic for
  textual inference}.
\newblock In \emph{Proceedings of the {ACL}-{PASCAL} Workshop on Textual
  Entailment and Paraphrasing}, pages 193--200, Prague. Association for
  Computational Linguistics.

\bibitem[{Minervini et~al.(2020)Minervini, Bosnjak, Rockt{\"{a}}schel, Riedel,
  and Grefenstette}]{Minerv}
Pasquale Minervini, Matko Bosnjak, Tim Rockt{\"{a}}schel, Sebastian Riedel, and
  Edward Grefenstette. 2020.
\newblock \href {https://doi.org/10.3233/SSW200015} {Differentiable reasoning
  on large knowledge bases and natural language}.
\newblock In Ilaria Tiddi, Freddy L{\'{e}}cu{\'{e}}, and Pascal Hitzler,
  editors, \emph{Knowledge Graphs for eXplainable Artificial Intelligence:
  Foundations, Applications and Challenges}, volume~47 of \emph{Studies on the
  Semantic Web}, pages 125--142. {IOS} Press.

\bibitem[{Nawaz et~al.(2019)Nawaz, Malik, Li, Sun, and Lali}]{nawaz2019survey}
M.~Saqib Nawaz, Moin Malik, Yi~Li, Meng Sun, and M.~Ikram~Ullah Lali. 2019.
\newblock \href {http://arxiv.org/abs/1912.03028} {A survey on theorem provers
  in formal methods}.

\bibitem[{Rockt\"{a}schel and Riedel(2017)}]{Rocktaschel2017}
Tim Rockt\"{a}schel and Sebastian Riedel. 2017.
\newblock End-to-end differentiable proving.
\newblock In \emph{Advances in Neural Information Processing Systems 30}, pages
  3788--3800.

\bibitem[{Russell and Norvig(2010)}]{russell2002artificial}
Stuart Russell and Peter Norvig. 2010.
\newblock \emph{Artificial intelligence: a modern approach}, 3 edition.
\newblock Prentice Hall.

\bibitem[{Saha et~al.(2020)Saha, Ghosh, Srivastava, and
  Bansal}]{saha2020prover}
Swarnadeep Saha, Sayan Ghosh, Shashank Srivastava, and Mohit Bansal. 2020.
\newblock \href {https://doi.org/10.18653/v1/2020.emnlp-main.9} {{PR}over:
  Proof generation for interpretable reasoning over rules}.
\newblock In \emph{Proceedings of the 2020 Conference on Empirical Methods in
  Natural Language Processing (EMNLP)}, pages 122--136, Online. Association for
  Computational Linguistics.

\bibitem[{Serafini and d'Avila Garcez(2016)}]{serafini2016logic}
Luciano Serafini and Artur~S. d'Avila Garcez. 2016.
\newblock Logic tensor networks: Deep learning and logical reasoning from data
  and knowledge.
\newblock In \emph{Proc. of the 11th International Workshop on Neural-Symbolic
  Learning and Reasoning (NeSy'16) co-located with the Joint Multi-Conference
  on Human-Level Artificial Intelligence {(HLAI} 2016)}, volume 1768.

\bibitem[{Singh et~al.(2020)Singh, Aggrawal, and Krishnamurthy}]{Singh2020}
Hrituraj Singh, Milan Aggrawal, and Balaji Krishnamurthy. 2020.
\newblock \href {http://arxiv.org/abs/2002.06544} {{Exploring Neural Models for
  Parsing Natural Language into First-Order Logic}}.

\bibitem[{Socher et~al.(2013)Socher, Chen, Manning, and Ng}]{NIPS2013_5028}
Richard Socher, Danqi Chen, Christopher~D. Manning, and Andrew~Y. Ng. 2013.
\newblock \href
  {https://proceedings.neurips.cc/paper/2013/hash/b337e84de8752b27eda3a12363109e80-Abstract.html}
  {Reasoning with neural tensor networks for knowledge base completion}.
\newblock In \emph{Advances in Neural Information Processing Systems 26: 27th
  Annual Conference on Neural Information Processing Systems 2013. Proceedings
  of a meeting held December 5-8, 2013, Lake Tahoe, Nevada, United States},
  pages 926--934.

\bibitem[{Sutcliffe et~al.(2004)Sutcliffe, Zimmer, and Schulz}]{data-exchange}
Geoff Sutcliffe, Jürgen Zimmer, and Stephan Schulz. 2004.
\newblock \emph{TSTP Data-Exchange Formats for Automated Theorem Proving
  Tools}, pages 201--215.

\bibitem[{Tafjord et~al.(2020)Tafjord, Mishra, and
  Clark}]{tafjord2020proofwriter}
Oyvind Tafjord, Bhavana~Dalvi Mishra, and Peter Clark. 2020.
\newblock Proofwriter: Generating implications, proofs, and abductive
  statements over natural language.
\newblock \emph{arXiv preprint arXiv:2012.13048}.

\bibitem[{Vaswani et~al.(2017)Vaswani, Shazeer, Parmar, Uszkoreit, Jones,
  Gomez, Kaiser, and Polosukhin}]{DBLP:journals/corr/VaswaniSPUJGKP17}
Ashish Vaswani, Noam Shazeer, Niki Parmar, Jakob Uszkoreit, Llion Jones,
  Aidan~N. Gomez, Lukasz Kaiser, and Illia Polosukhin. 2017.
\newblock \href
  {https://proceedings.neurips.cc/paper/2017/hash/3f5ee243547dee91fbd053c1c4a845aa-Abstract.html}
  {Attention is all you need}.
\newblock In \emph{Advances in Neural Information Processing Systems 30: Annual
  Conference on Neural Information Processing Systems 2017, December 4-9, 2017,
  Long Beach, CA, {USA}}, pages 5998--6008.

\bibitem[{Vovk(2015)}]{DBLP:journals/corr/Vovk15}
Vladimir Vovk. 2015.
\newblock \href {http://arxiv.org/abs/1502.06254} {The fundamental nature of
  the log loss function}.
\newblock \emph{CoRR}, abs/1502.06254.

\bibitem[{Weber et~al.(2019)Weber, Minervini, M{\"u}nchmeyer, Leser, and
  Rockt{\"a}schel}]{Weber2019NLPrologRW}
Leon Weber, Pasquale Minervini, Jannes M{\"u}nchmeyer, Ulf Leser, and Tim
  Rockt{\"a}schel. 2019.
\newblock \href {https://doi.org/10.18653/v1/p19-1618} {Nlprolog: Reasoning
  with weak unification for question answering in natural language}.
\newblock In \emph{Proc. of 57th Conference of the Association for
  Computational Linguistics, {ACL}}, volume~1, pages 6151--6161.

\bibitem[{Zellers et~al.(2018)Zellers, Bisk, Schwartz, and
  Choi}]{zellers2018swagaf}
Rowan Zellers, Yonatan Bisk, Roy Schwartz, and Yejin Choi. 2018.
\newblock \href {https://doi.org/10.18653/v1/D18-1009} {{SWAG}: A large-scale
  adversarial dataset for grounded commonsense inference}.
\newblock In \emph{Proceedings of the 2018 Conference on Empirical Methods in
  Natural Language Processing}, pages 93--104, Brussels, Belgium. Association
  for Computational Linguistics.

\end{thebibliography}
\bibliographystyle{acl_natbib}


\clearpage

\appendix
\section{Input/Output of the Neural Unifier's units}

This section provides a detailed example of the input and output of the core units of the architecture.

\subsection{Neural fact checking unit}

\subsubsection{Training phase}

\begin{itemize}[leftmargin=*]

\item Input: $(\kappa, q_0)$,

where $\kappa$ is the set of facts and rules concatenated in a single string and $q_0$ is the depth-0 query, for example:
$$\kappa = \text{"Bob is big.Gary is not cold. ..."}$$
$$q_0 = \text{"Bob is big?"}$$

In the training phase, the input of the fact checking unit is furthermore tokenized and transformed into a numerical vector (using the BERT embedding layer) that follows the format:
$[CLS] \: C \: [SEP] \: Q_0 \: [SEP]$, 

where $C$ is the embedding of $\kappa$ and $Q_0$ is the embedding of $q_0$.
$[CLS]$ and $[SEP]$ are embedding vectors of special tokens added by the BERT tokenizer to separate context and query \cite{bert}.

\item Output: $(True/False)$

\end{itemize}

\subsubsection{Inference phase}

When the fact checking unit is used for training the neural unification unit or in the inference phase (both with frozen weights), the input skips the tokenization phase.

\begin{itemize}[leftmargin=*]

\item Input: $[CLS] \: C \: [SEP] \: NU_O \: [SEP]$,

where $C$ is the embedding of $\kappa$ and $NU_O$ is the vector embedding given in output by Neural unification unit.

\item Output: $(True/False)$

\end{itemize}

\subsection{Neural unification unit}

\begin{itemize}[leftmargin=*]

\item Input: $(\kappa, q_n)$,

As for the other unit, the input of the unification unit is furthermore tokenized and transformed in the corresponding BERT embeddings, following the format: $[CLS] \: C \: [SEP] \: Q_n \: [SEP]$, 

where $C$ is the embedding of $\kappa$ and $Q_n$ is the embedding of $q_n$.
$[CLS]$ and $[SEP]$ are embedding vectors of special tokens added by the BERT tokenizer to separate context and query \cite{bert}.

\item Output: $[CLS] \: C_0 \: [SEP] \: NU_O \: [SEP]$,

where $C_0$ is the embedding vector in output corresponding to the tokens of the given input context.

As explained in the paper, $C_0$ is replaced with $C$ in the embedding, before fed it to the fact checking unit.

\end{itemize}

\section{Datasets}

Tables \ref{distribution-train-f1}, \ref{distribution-train-f2}, \ref{distribution-test-par} and \ref{distribution-test-electric} report some additional statistics of the three datasets used. Detailed examples of dataset instances are shown in \cite{clark2020transformers}.

\begin{table}[h!]
\caption{Distribution of CWA  queries in the train data in folder 1 (F=1).}
\label{distribution-train-f1}
\vskip -0.10in
\begin{center}
\begin{small}
\begin{sc}
\center{\includegraphics[width=.49\textwidth]
{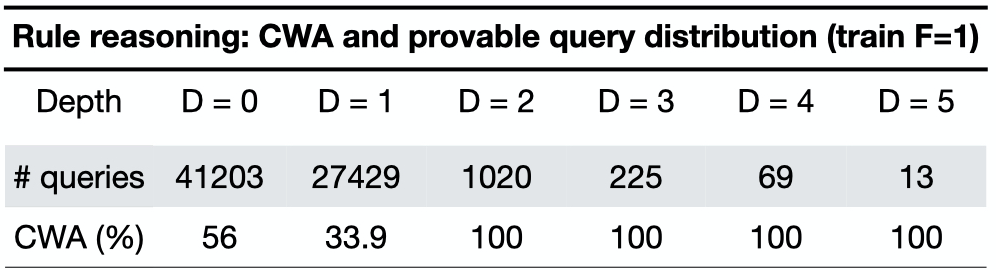}}
\end{sc}
\end{small}
\end{center}
\vskip -0.30in
\end{table}

\begin{table}[h!]
\caption{Distribution of CWA  queries in the train data in folder 2 (F=2).}
\label{distribution-train-f2}
\vskip -0.10in
\begin{center}
\begin{small}
\begin{sc}
\center{\includegraphics[width=.49\textwidth]
{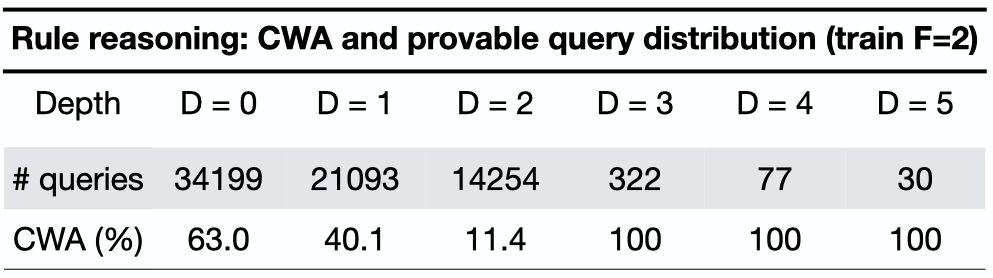}}
\end{sc}
\end{small}
\end{center}
\vskip -0.30in
\end{table}

\begin{table}[h!]
\caption{Distribution of CWA queries in the Paraphrased test data.}
\label{distribution-test-par}
\vskip -0.10in
\begin{center}
\begin{small}
\begin{sc}
\center{\includegraphics[width=.49\textwidth]
{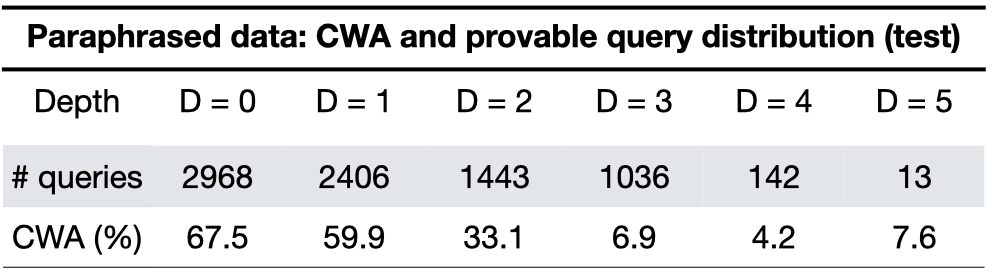}}
\end{sc}
\end{small}
\end{center}
\vskip -0.30in
\end{table}

\begin{table}[h!]
\caption{Distribution of CWA queries in the Electricity test data.}
\label{distribution-test-electric}
\vskip -0.10in
\begin{center}
\begin{small}
\begin{sc}
\center{\includegraphics[width=.49\textwidth]
{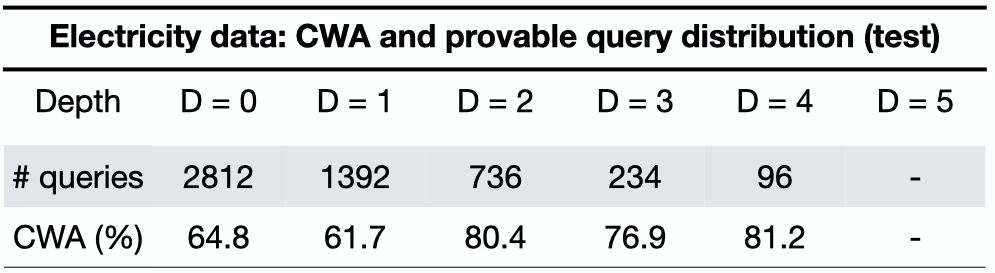}}
\end{sc}
\end{small}
\end{center}
\vskip -0.30in
\end{table}

\section{Results with different types of transformers}

Besides BERT, we also tried to use ROBERTA and BERT fine-tuned with the scale adversarial dataset for grounded commonsense inference \cite{zellers2018swagaf}. The results are illustrated in Table \ref{tb:methods}, which shows that our results are not specific to BERT, but instead our approach works well also for other types of models.

\begin{table}[h!]
\caption{\label{tb:methods} Results on the rule reasoning and the paraphrased datasets with different types of transformers used as the basis of the NU neural network}
\vskip -0.10in
\begin{center}
\begin{small}
\begin{sc}
\center{\includegraphics[width=.49\textwidth]
{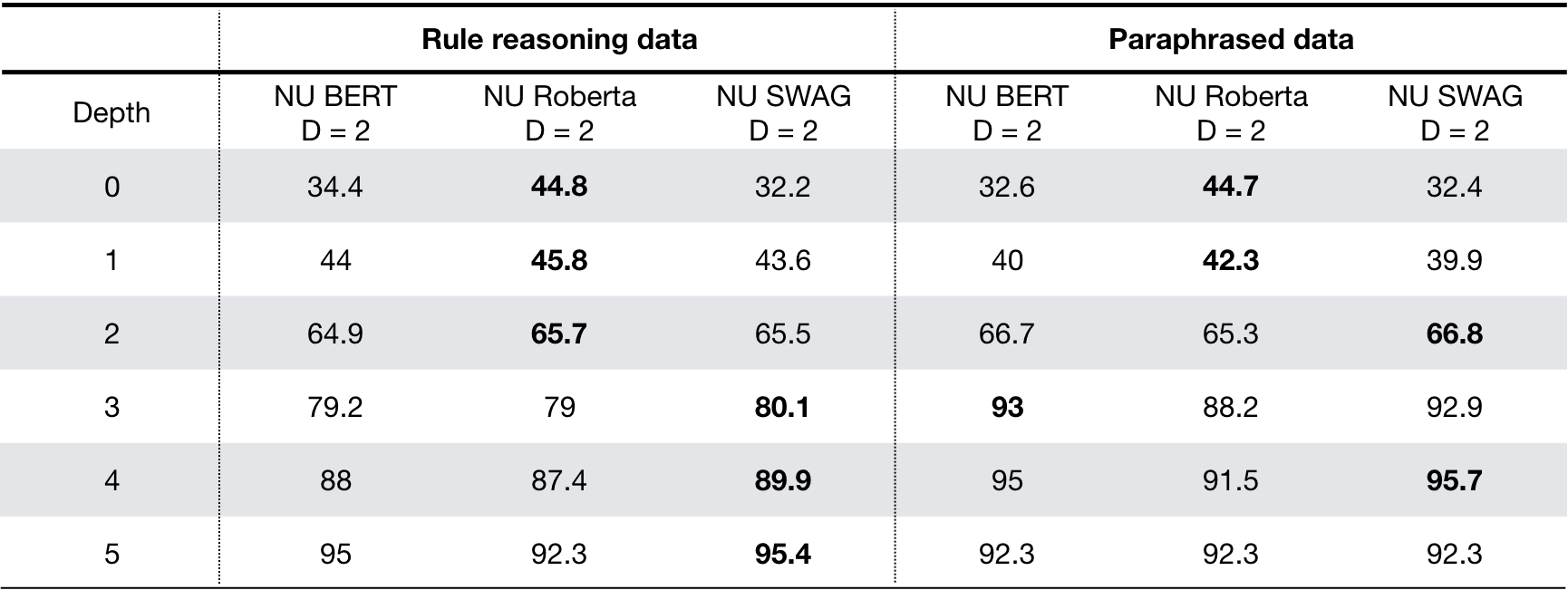}}
\end{sc}
\end{small}
\end{center}
\vskip -0.30in
\end{table}

\section{Runtime information and computing infrastructure}

The experiments reported in the paper were performed on a cloud cluster with a Tesla v100 GPU, 16 GB of RAM and SSD. The training of the fact checking unit on this instance takes less than 60 minutes, while the training of the unification unit takes less than 240 minutes (4 hours). The inference times are less than two minutes for all datasets.


\end{document}